\documentclass{article}
\usepackage{ijcai21}

\usepackage{times}
\usepackage{soul}
\usepackage{url}
\usepackage[hidelinks]{hyperref}
\usepackage[utf8]{inputenc}
\usepackage[small]{caption}
\usepackage{graphicx}
\usepackage{amsmath}
\usepackage{amsthm}
\usepackage{booktabs}
\usepackage{algorithm}
\usepackage{algorithmic}
\usepackage{booktabs}
\usepackage{tabularx}
\usepackage[flushleft]{threeparttable}
\usepackage{multirow}
\usepackage{comment}
\usepackage{tikz}
\usepackage{colortbl}
\usepackage{subcaption}
\usepackage{bm}
\usepackage{epigraph}

\definecolor{highlight}{HTML}{ffeecc}
\definecolor{Gray}{HTML}{f2f2f2}

\newcommand{\articles}[0]{\textsc{AI/ML~News}}
\newcommand{\abstracts}[0]{\textsc{AI/ML~Abstracts}}

\title{A Large-Scale, Automated Study of Language Surrounding Artificial Intelligence}

\author{
    Autumn Toney
    \affiliations
    Center for Security and Emerging Technology \\ Georgetown University
    \emails
    autumn.toney@georgetown.edu
}

\begin{document}

\maketitle

\begin{abstract}

% an idea for changing up the presentation of the research

This work presents a large-scale analysis of artificial intelligence (AI) and machine learning (ML) references within news articles and scientific publications between 2011 and 2019. We implement word association measurements that automatically identify shifts in language co-occurring with AI/ML and quantify the strength of these word associations. Our results highlight the evolution of perceptions and definitions around AI/ML and detect emerging application areas, models, and systems (e.g., \textit{blockchain} and \textit{cybersecurity}). Recent small-scale, manual studies have explored AI/ML discourse within the general public, the policymaker community, and researcher community, but are limited in their scalability and longevity. Our methods provide new views into public perceptions and subject-area expert discussions of AI/ML and greatly exceed the explanative power of prior work. 
\end{abstract}

\section{Introduction}
\epigraph{You shall know a word by the company it keeps.}{J. R. Firth}

Referenced in a wide-range of domains, from science fiction to autonomous vehicles, artificial intelligence (AI) has gained significant attention from societies and governments worldwide. Despite its emerging prominence in the public sphere, AI still lacks a consistent, universally-accepted definition, making it a challenging subject to analyze over time \cite{bryson2019past,cave2018portrayals,chuan2019framing,fast2017,krafft2020defining,legg2007universal}. Recent studies conducted surveys and manual annotation tasks to understand how subject-area experts, the general population, and policy makers define and perceive AI \cite{fast2017,krafft2020defining,cave2018portrayals,cave2019scary,chuan2019framing,russell2002artificial,sweeney2003s}. While these studies are a necessary preliminary step in uncovering historical and current perceptions of AI, these studies are limited in scalability and constrained to the period of time in which they were performed.

To improve on these limitations and gain new insights, we present a large-scale, automated approach to analyze the language surrounding AI in the public sphere during any time period. Using text corpora from news articles and scientific publication abstracts, we analyze AI references in two domains: one that represents public perceptions of AI and one that represents subject-area expert applications of AI. Our analysis includes more than 170,000 AI-related news articles and 77,000 AI-related scientific publication abstracts. To the best of our knowledge, our work is the largest-scale study on AI references in the public sphere, specifically in text corpora.

Our approach uses word association structures in text corpora and measures the strength of and the shifts in word associations over time. In psycholinguistics, word association structures are often identified through human studies where participants are presented with a word (e.g., \textit{coffee}) and respond with a word that comes to mind (e.g., \textit{mug}). Word association structures in text corpora can be automatically identified by analyzing words that frequently co-occur within a designated proximity of each other. Thus, word association structures can be automatically derived from a text corpus, indicating the different characteristics of a word by the ``company it keeps.'' %Additionally, word association structures can be identified in text corpora regardless of the time period, providing a dynamic measurement for language surrounding a topic of interest. 

We use mutual information to measure the strength of association and a normalized co-occurrence frequency value to measure the shifts in frequently co-occurring words over time. Mutual information is an indicator of words that have a high probability of exclusively co-occurring with a target word. In this way, mutual information values identify words that have an exclusive co-occurrence with a target word, as opposed to words with a general co-occurrence. For example, in news articles, we find that \textit{robotics} has a high co-occurrence frequency with \textit{artificial intelligence} and \textit{machine learning} and a high mutual information value, whereas \textit{big} has a high co-occurrence frequency but a low mutual information value. Shifts in co-occurrence frequency ranks indicate words that are emerging, decreasing in frequency, or increasing in frequency. For example, in scientific publication abstracts, we find that \textit{convolutional} is an emerging word co-occurring with \textit{artificial intelligence} and \textit{machine learning}.

In the following sections, we provide a background on word association structures (Section 2), summarize related work (Section 3), describe the datasets studied in our analysis (Section 4), define our methodology (Section 5), and present and discuss our experimental results (Sections 6 and 7).

\section{Background}
In psychology, the law of mental association defines the phenomena of learning by contiguity, a learning process that associates a stimulus and response based on their frequency and proximity (e.g., coffee being associated with mug) \cite{james2007principles}. Applied to linguistic theory, the law of mental association relates to language acquisition; words associated to a particular concept are stored closely in a human's ``mental lexicon'' \cite{dobel2010new}. When words frequently co-occur, by some definition of proximity, their association in a mental lexicon is strengthened \cite{wettler1993computation,church1990word}.    

Word associations are dynamic, as language evolves associations will change \cite{nelson2004university}. Prior psycholinguistic studies identify word association norms across populations \cite{nelson2004university,wettler1993computation,church1990word,buchanan2019english}. These studies commonly use priming---showing a stimulus (an image or word)---and measure the speed of a response or the consistency of responses across the participants. For example, Nelson et al. conducted a free response survey where participants were asked to write the first word that came to mind after reading a cue word \cite{nelson2004university}. This survey was designed to capture associative knowledge and characteristics of meaning; responses were shown to be affected by culture and trends. Consistent word associations across participants indicate a common experience with words, and inconsistent word associations across participants highlight experiences that vary from the norm \cite{nelson2004university}. 

%In this experiment, the stimulus was the cue word and the participants' responses were unconstrained.

%To replicate human survey results and derive local information of word associations, word co-occurrence is defined as two words appearing within a designated window size of each other \cite{gunther2016predicting}.

These human surveys are translated to automated procedures performed on text corpora, providing a scalable analysis of word associations, by defining word co-occurrences as two words appearing within a designated window size of each other \cite{gunther2016predicting}. Window size defines a proximity constraint for word co-occurrence; for example, a window size of two considers only two words to the left and two words to the right of the target word. Wettler and Rapp find that a window size of five is optimal for large text corpora, as it does not dilute the language surrounding a target word and maintains a close enough proximity to capture true association \cite{wettler1993computation}.

 %If co-occurrence is defined as co-occurring in the same document, one can derive global lexical information, whereas if co-occurrence is defined as co-occurring in a window size, one can derive local lexical information.

Word co-occurrences, measured by using a specified window size, have been used in natural language processing tasks, such as generating semantic spaces \cite{lund1996producing}. In practice, applying word association methods on large-scale text corpora eliminates the sample bias of participants, as participant judgements are used to measure norms. However, word association methods do not eliminate other types of biases captured in linguistic norms, though they have also proven useful in this space \cite{caliskan2017semantics,bolukbasi2016man}.   
\section{Related Work}

Previous studies have taken various manual approaches to define AI and present public perceptions of AI. Russell and Norvig analyzed AI defintions in eight textbooks published between 1978 and 1993, and then specified four main ways AI is defined: 1) think like humans, 2) act like humans, 3) think rationally, and 4) act rationally \cite{russell2002artificial}. Building on Russell and Norvig's work, Sweeney manually categorized 996 AI-related publications cited by Russell and Norvig \cite{sweeney2003s}. Sweeney found that 987 of these publications favor defining AI in terms of rational thinking and rational behavior \cite{sweeney2003s}. 

Cave et al. surveyed 1,078 UK participants and collected responses from multiple choice and free response questions to learn about public perceptions of AI \cite{cave2019scary}. Notably, 85\% of respondents stated that they had heard of AI before, with 25\% of them defining AI in terms of robots. Krafft et al. conducted two surveys, one with 98 participants and one with 86 participants, where the authors asked AI researchers what they consider AI systems to be and how they define AI in practice \cite{krafft2020defining}. They compared the survey responses to policy definitions of AI, which they collected by manually annotating 83 policy documents from 2017 through 2019 \cite{krafft2020defining}. Krafft et al. found that policy documents typically use ``human-like'' definitions of AI, wheres AI researchers define AI through technical problems and functionality \cite{krafft2020defining}.

%Other results of this study showcased that, in general, the respondents felt businesses, research, and the government were in control of AI and its development

Fast and Horvitz analyzed AI-related news articles from the New York Times between 1986 and 2016, approximately 3 million articles in total \cite{fast2017}. Any paragraph in an article that mentioned the terms \textit{artificial intelligence}, \textit{AI}, or \textit{robot} was selected, reducing the data down to 8,000 paragraphs over the thirty years. The paragraphs were manually annotated by Amazon Mechanical Turkers, and the results describe trends in the public perception of AI over time. Specifically, mentions of AI have increased, the general population has become more optimisitc about AI, and concerns over the loss of control of AI are increasing \cite{fast2017}. Chuan et al. sampled news articles from LexisNexis and ProQuest from five U.S. news sources (USA Today, The New York Times, Los Angeles Times, New York Post, and Washington Post) that contain the term \textit{artificial intelligence}. Using stratified sampling, they reduced the 2,485 AI-related articles to 399 articles that are manually annotated by three graduate students. Chuan et al.'s study focused more on understanding the framing of AI in news articles and presented findings on the main topics, cited sources, and sentiment in their subset of AI-related news articles. They found that AI was mainly discussed in \textit{Business and Economy} and \textit{Science and Technology} article topics and that AI ethics is increasingly discussed \cite{chuan2019framing}.

\section{Datasets}

We study two large-scale datasets to generate subsets of AI/ML text data: 1) \textbf{\articles{}}, 170,858 news articles from the LexisNexis database \cite{LN} and 2) \textbf{\abstracts{}}, 77,880 scientific publication abstracts from the Microsoft Academic Graph \cite{sinha2015overview}. We categorize an article or abstract as AI/ML if it contains the terms \textit{artificial intelligence} or \textit{machine learning} at least once, using Bryson's description of important terms for understanding AI \cite{bryson2019past}. For both news articles and scientific publication abstracts, we normalize the text by setting all words to lower case and removing symbols, digits, URLs, email addresses, phone numbers, and punctuation except for apostrophes.  Additionally, we remove all stop words using NLTK's English set of stop words.\footnote{\url{https://www.nltk.org/}}

%The final sets of news articles (\articles{}) and scientific publication abstracts (\abstracts{}) used in our analysis contain the cleaned text. 

\textbf{LexisNexis Database:}
The LexisNexis database contains news article texts that were published between 2011 and 2020. We analyze English-language articles from 2011, 2015, and 2019 that were published by sources of good editorial quality. LexisNexis generates source editorial rankings for news articles on a rank scale is from 1 to 5, with 1 being high quality (e.g., The New York Times) and 5 being low quality (e.g., message boards). We select news articles that have an editorial rank of 1, 2, or 3, which includes international, national, business, regional, industry, and government news sources. We use the duplicate ID assigned by LexisNexis to de-duplicate the articles. The 170,858 news articles in \articles{} is comprised of these filtered and de-duplicated documents.
%\articles{} contains the cleaned and filtered text.

\begin{figure}[th]
    \centering
       \includegraphics[width=0.48\textwidth]{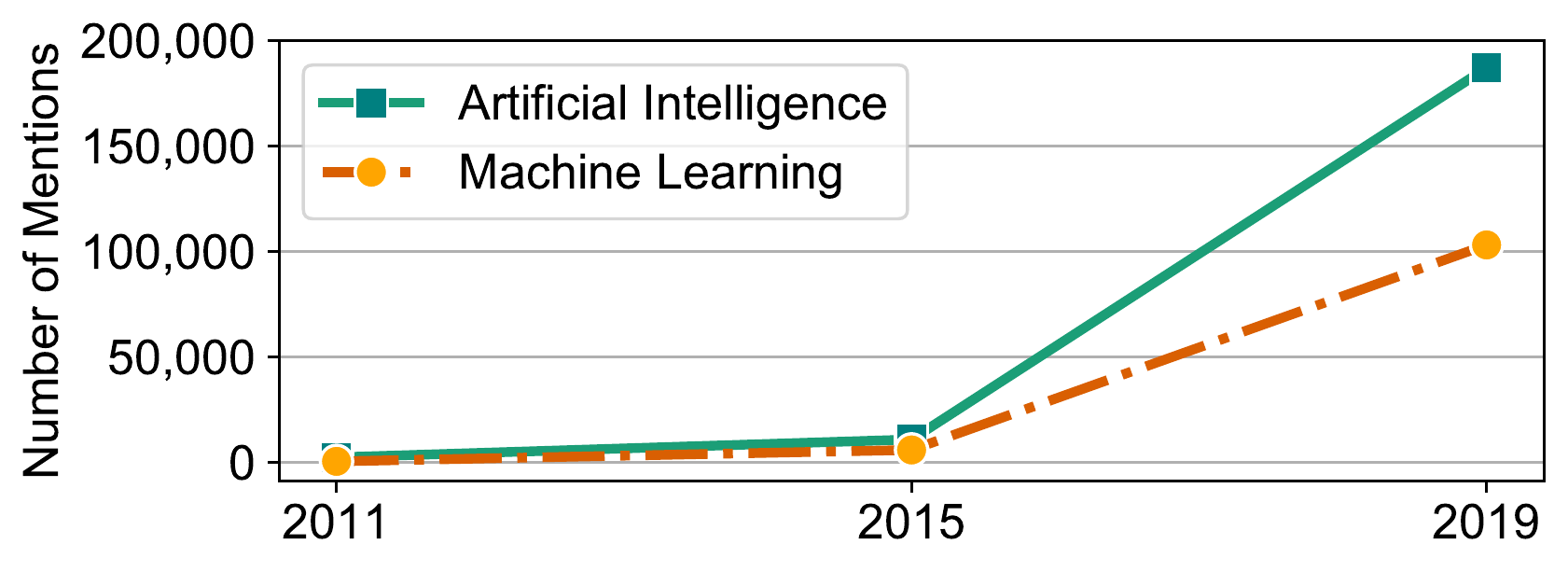}
        \caption{Term counts for \textit{artificial intelligence} and \textit{machine learning} respectively in \articles{} over time}
    \label{fig:topic mentions}
\end{figure}

Table \ref{tab:LN_description} provides details for each year's subset of \articles{}. Over time, the number of documents, tokens (unique vocabulary words), and sources significantly increase. Figure \ref{fig:topic mentions} displays the counts of \textit{artificial intelligence} and \textit{machine learning} mentions in \articles{}. Mentions of artificial intelligence are more frequent than mentions of machine learning over the entire period of study; there are 2,446 AI mentions and 554 ML mentions in 2011 and 187,066 AI mentions and 103,175 ML mentions in 2019. 

%Documents in \articles{} have average word counts of $1,270$ in 2011, $1,078$ in 2015, and $1,290$ in 2019. Additionally, \articles{} has increasing numbers of sources ($451$ in 2011, $932$ in 2015, and $3,042$ in 2019) and increasing numbers of news articles ($2,143$ in 2011, $10,345$ in 2015, and $158,370$ in 2019) over time. 

\newcolumntype{P}{>{\raggedleft\arraybackslash}p{.64in}}
\newcolumntype{O}{>{\raggedleft\arraybackslash}p{.7in}}
\newcolumntype{M}{>{\raggedleft\arraybackslash}p{.45in}}
\newcolumntype{n}{>{\raggedleft\arraybackslash}p{.5in}}

\begin{table}[h]\small
    \centering
    \begin{tabular}{p{.15in}OPnM}
    \toprule
         \textbf{Year} & \textbf{Num. of Documents} & \textbf{Avg. Word Count} &\textbf{Num. of Tokens} & \textbf{Num. of Sources} \\
         \midrule
         2011 & 2,143 & 1,270 & 76,182 & 451\\
         2015 & 10,345 & 1,078 & 163,604 & 932\\
         2019 & 158,370 & 1,290 & 1,021,275 & 3,042\\
        \bottomrule

    \end{tabular}
    \caption{Details of the \articles{} corpus}
    \label{tab:LN_description}
\end{table}
\vspace{-7mm}

\textbf{Microsoft Academic Graph:}
Microsoft Academic Graph (MAG) contains scientific research publication documents from eight categories: Book, Book Chapter, Conference, Dataset, Journal, Patent, Repository, and Thesis \cite{sinha2015overview}. We use a subset of MAG documents from 2011, 2015, and 2019 that contain an abstract in their publication record. 

%Documents in \abstracts{} have average word counts of $180$ in 2011, $187$ in 2015, and $182$ in 2019, which is comparatively fewer words per text instance than \articles{}. There is an increasing number of \abstracts{} over time, with $7,693$ in 2011, $13,432$ in 2015, and $56,755$ in 2019, which is comparatively fewer text instances than in \articles{}. 

\begin{figure}[th]
    \centering
       \includegraphics[width=0.48\textwidth]{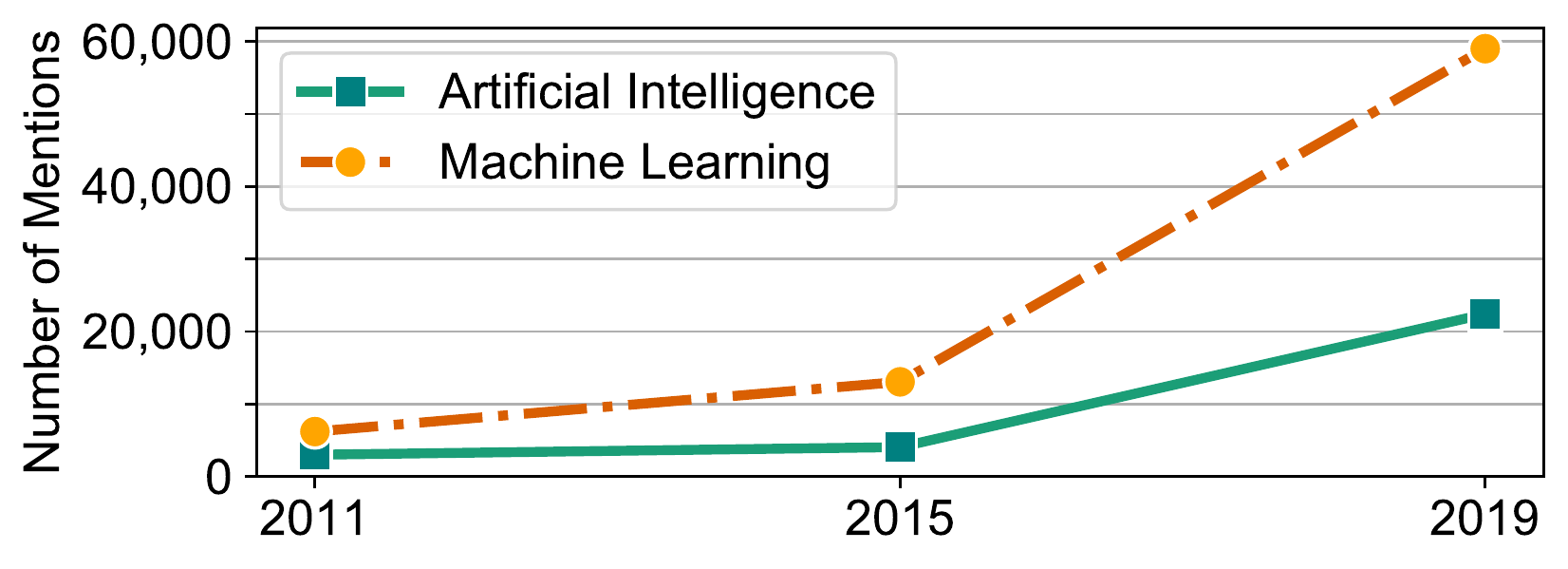}
        \caption{Term counts for \textit{artificial intelligence} and \textit{machine learning} respectively in \abstracts{} over time}
    \label{fig:topic mentions abstracts}
\end{figure}

Table \ref{tab:MAG_description} provides details for each year's subset of \abstracts{}. There are comparatively fewer words per text instance and fewer documents in \abstracts{} than in \articles{}. Figure \ref{fig:topic mentions abstracts} displays the counts of \textit{artificial intelligence} and \textit{machine learning} mentions in \abstracts{}. Mentions of machine learning are more frequent than mentions of artificial intelligence over the entire period of study, the opposite of AI/ML mentions in \articles{}. In 2011, there are 6,210 ML mentions and 3,012 AI mentions, and in 2019, there are 59,006 ML mentions and 22,414 AI mentions.

\begin{table}[h]\small
    \centering
    \begin{tabular}{cOPMM}
    \toprule
         \textbf{Year} & \textbf{Num. of Documents} & \textbf{Avg. Word Count} &\textbf{Num. of Tokens} \\
         \midrule
         2011 & 7,693 & 180 & 48,449\\
         2015 & 13,432 & 187 & 69,624\\
         2019 & 56,755 & 182 & 158,097 \\
         \bottomrule
    \end{tabular}
    \caption{Details of the \abstracts{} corpus}
    \label{tab:MAG_description}
\end{table}
\vspace{-7mm}
\vspace{-5mm}
\section{Methodology}
\begin{figure}
\centering
\input{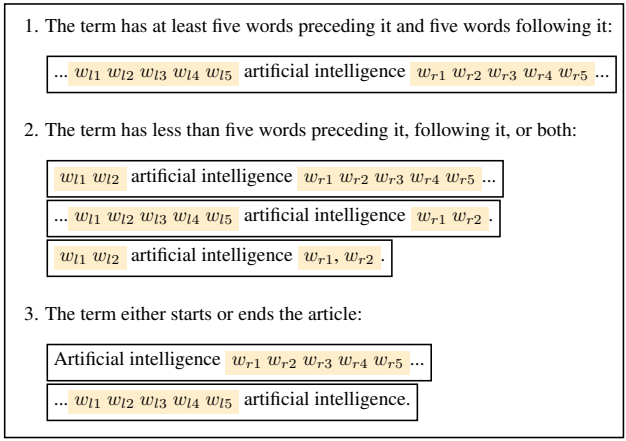}
\caption{Examples of co-occurring words within a designated window size of five to \textit{artificial intelligence} (AI). The yellow highlight indicates words considered to be co-occurring with AI.}
\label{fig:windows}
\end{figure}

%reword
We use two word association measurements to provide a comprehensive understanding of how words co-occurring with \textit{artificial intelligence} and \textit{machine learning} change over time: mutual information and normalized co-occurrence rank. Both measurements rely on a definition of co-occurrence, thus we define co-occurrence as a word co-occurring within a window size of the terms \textit{artificial intelligence} and \textit{machine learning}. Since AI and ML are two-word terms, we consider words to the left of \textit{artificial/machine} and words to the right of \textit{intelligence/learning} within the defined window size. We account for edge cases in selecting co-occurring words, such as \textit{artificial intelligence} or \textit{machine learning} ending a document. Figure \ref{fig:windows} demonstrates term co-occurrences within a five-word window under various text positions. 

%2011: 5791 AI mentions, 1819 ML mentions
%2015: 25000 AI mentions, 14298 ML mentions
%2019: 299758 AI mentions, 171683 ML mentions

%HIGH QUALITY
%2011: 2446 AI mentions 554 ML mentions
%2015: 10916 AI mentions 5957 ML mentions
%2019: 187066 AI mentions 103175 ML mentions

\subsection{Mutual Information}

We define mutual information between two words according to Church and Hanks \cite{church1990word}. Given two words, $w_1$ and $w_2$, their mutual information $I(w_{1}, w_{2})$ is defined as:
\begin{align}
I(w_{1}, w_{2}) = \log_{2}\frac{\Pr[W_{1}=w_{1}, W_{2}=w_{2}]}{\Pr[W_{1}=w_{1}]\Pr[W_{2}=w_{2}]}
\end{align}

$\Pr[W=w]$ is the probability that a word drawn at random from a document in the text corpus is equal to $w$. Specifically, $\Pr[W=w] = \frac{w_{count}}{total_{count}}$, where $w_{count}$ is the number of times that $w$ appears in the document and $total_{count}$ is the number of words in the document. $\Pr[W_{1}=w_{1}, W_{2}=w_{2}]$ is the joint probability that the two words co-occur (within a window size) in a text corpus, indicating association. If two words frequently co-occur in text, $\Pr[W_{1}=w_{1}, W_{2}=w_{2}]$ will be a larger value than if they infrequently co-occur; thus, a stronger association between two words (frequent co-occurrences) results in a larger value for $I(w_{1}, w_{2})$. In this way, mutual information quantifies the \textit{strength of association} between word co-occurrences and provides a metric that can be compared over time. 

%$\Pr(w_{1})\Pr(w_{2})$ is the probability of $w_1$ and $w_2$ appearing in a text corpus independently of each other, indicating chance.

\subsection{Normalized Co-occurrence Rank}  
In order to identify words that emerge or are increasing or decreasing in their frequency of co-occurrence over time, we define a normalized co-occurrence rank value. Normalization is necessary for frequency rank comparisons over time, since there is a significant increase of tokens from year to year in both \articles{} and \abstracts{} (see Table \ref{tab:LN_description} and \ref{tab:MAG_description}). We identify the set of co-occurring words within the designated window size and count their frequency of co-occurrence with either term (\textit{artificial intelligence} or \textit{machine learning}) in a given year. We sort the co-occurring frequency word set in descending order of frequency. This ordering assigns a rank value to each word, with the most frequently co-occurring word at rank $1$. We compute the normalized rank by dividing the assigned ranks by the total number of words in the co-occurring frequency word set in a given year. For a word $w$ and a year $y$, where $rank_{w}$ is the assigned rank of the word $w$ and $total_{y}$ is the number of words in the year's co-occurring frequency word set, the normalized rank $N(w,y)$ is defined as:
\begin{align}
N(w,y) = \frac{rank_{w}}{total_{y}}
\end{align}

%For example, a word $w$ in \articles{} in a given year's co-occurring frequency word set, the normalized rank ($N(w,y)$) is defined as $N(w,y) = \frac{rank_{w}}{total_{y}}$, where $rank_{w}$ is the assigned rank of the word $w$ and $total_{y}$ is the number of words in the year's \articles{} co-occurring frequency word set. We round the normalized ranks to the nearest 0.05 to smooth the results.  

We compute the normalized ranks of co-occurring words for each year (2011, 2015, and 2019), rounding the normalized ranks to the nearest 0.05 to smooth the results. Then we compute the standard deviation ($\sigma$) of normalized ranks for each word over the three years. Words with lower $\sigma$ values shift minimally from year to year, indicating words that maintain their co-occurrence frequency over time. Words with higher $\sigma$ values shift maximally from year to year, indicating words that emerge or have a downward or upward co-occurrence frequency trend over time. In this way, normalized co-occurrence rank identifies the \textit{shift of language} over time.

\begin{comment}
For a given AI/ML article or abstract, we iterate through its words to identify the pair of indices for a term of interest, $(idx_{1}, idx_{2})$, where $idx_1$ corresponds to the location of ``artificial''/``machine'' and $idx_2$ corresponds to the location of ``intelligence''/``learning''. 

We compute the total co-occurring frequencies for all \articles{} and \abstracts{}, and sort our finalized list in descending order of frequency.
\end{comment}
\section{Experiments and Results}
\begin{comment}

Our first step in experimentation is to compute the raw word frequencies for all unique words in our sets of articles snd abstracts for each year. Table \ref{tab:tokens} shows the increase in unique words observed in the article text over time. We find that AI/ML articles have a significantly large increase in vocabulary over time compared to AI/ML abstracts, with a 1,431\% increase from 2015 to 2019 compared to 127\% increase. 
\end{comment}
\begin{comment}

\begin{table}[th]
    \centering
        \begin{tabular}{c c r r}
        \toprule
        \textbf{Type} & \textbf{Year} & \textbf{Num. Tokens} & \textbf{Increase} \\ 
        \midrule
        {\multirow{3}{*}{Articles}}& 2011 & 2,143 & -- \\

        & 2015 & 10,345 & 383\% \\
        & 2019 & 158,370 & 1431\% \\ 
        \\
        {\multirow{3}{*}{Abstracts}} & 2011 & 48,449 & -- \\ 

        & 2015 & 69,624  & 44\% \\ 
        & 2019 & 158,097 & 127\%\\ 

        \bottomrule
        \end{tabular}
        \caption{The number of unique words present in each set of AI/ML articles and abstracts over three years and their percent increase.}
        \label{tab:tokens}
\end{table}

\end{comment}

Our first step in experimentation is to compute the word co-occurrence frequencies, using a window size of five as recommended by Wettler and Rapp \cite{wettler1993computation}, for each year respectively. We also tested windows with sizes three and eight in our experiments, but found that the results vary minimally, thus we present the results for window size of five (see Supplementary Materials for details). The computed word co-occurrence frequencies provides us with the necessary data to apply our word association measurements. 

For each year's results, we sort the words by descending order of their frequency. Table \ref{tab: timeline} and \ref{tab: timeline abstracts} display the top 15 most frequently co-occurring words with AI/ML over time in \articles{} and \abstracts{} respectively. We find that \abstracts{} have a more consistent set of top 15 co-occurring words over time, with minimal words being introduced or being dropped in each year, compared to the top 15 co-occurring words in \articles{}. In \articles{}, four out of the 15 words consistently appear, such as \textit{technology} and \textit{data}. In \abstracts{}, 10 out of the 15 words consistently appear, such as \textit{techniques} and \textit{algorithms}. We find that \textbf{\textit{data}} and \textbf{\textit{using}} appear in both \articles{} and \abstracts{} for all three years.

%Comparing the top 15 most frequently co-occurring words across \articles{} and \abstracts{} in all years, we see \textbf{\textit{data}} and \textbf{\textit{using}} appear in both sets for all years.  

\newcommand{\foo}{\makebox[0pt]{\textbullet}\hskip-0.5pt\vrule width 1pt\hspace{\labelsep}}

\begin{table*}[th]\small
\centering
\begin{subtable}{0.5\textwidth}
\renewcommand\arraystretch{1.4}
%\captionsetup{singlelinecheck=false, font=blue, labelfont=sc, labelsep=quad}
%\caption{Articles}\vskip -1.5ex
\hspace{-5mm}
\begin{tabular}{@{\,}r <{\hskip 2pt} !{\foo} >{\raggedright\arraybackslash}p{7.5cm}}
%\toprule
\addlinespace[1.5ex]
2011 & computer, \textcolor{blue}{\textbf{technology}}, \textcolor{blue}{\textbf{ai}}, science, software, research, \textcolor{blue}{\textbf{data}}, techniques, \textcolor{blue}{\textbf{using}}, uses, use, algorithms, robotics, said \\

2015 & \textcolor{blue}{\textbf{data}}, \textcolor{blue}{\textbf{technology}}, \textcolor{blue}{\textbf{ai}}, analytics, big, new, \textcolor{blue}{\textbf{using}}, computer, technologies, science, research, said, algorithms, robotics, also, human \\

2019 & \textcolor{blue}{\textbf{ai}}, \textcolor{blue}{\textbf{data}}, \textcolor{blue}{\textbf{technology}}, technologies, intelligence, analytics, artificial, \textcolor{blue}{\textbf{using}}, new, use, big, digital, learning, company, internet \\
 
\end{tabular}
\caption{\articles{}}  \vskip -1.5ex
\label{tab: timeline}
\end{subtable}
\hspace{-1mm}
\begin{subtable}{0.45\textwidth}
\renewcommand\arraystretch{1.4}
\begin{tabular}{@{\,}r <{\hskip 2pt} !{\foo} >{\raggedright\arraybackslash}p{7.3cm}}
\addlinespace[1.5ex]

2011 & \textcolor{blue}{\textbf{techniques}}, \textcolor{blue}{\textbf{data}}, \textcolor{blue}{\textbf{methods}}, \textcolor{blue}{\textbf{based}}, \textcolor{blue}{\textbf{using}}, \textcolor{blue}{\textbf{algorithms}}, \textcolor{blue}{\textbf{used}}, \textcolor{blue}{\textbf{method}}, paper, \textcolor{blue}{\textbf{learning}}, approach, \textcolor{blue}{\textbf{system}}, research, mining, classification \\

2015 & \textcolor{blue}{\textbf{data}}, \textcolor{blue}{\textbf{techniques}}, \textcolor{blue}{\textbf{algorithms}}, \textcolor{blue}{\textbf{using}}, \textcolor{blue}{\textbf{methods}}, \textcolor{blue}{\textbf{based}}, \textcolor{blue}{\textbf{used}}, approach, algorithm, \textcolor{blue}{\textbf{method}}, classification, paper, \textcolor{blue}{\textbf{learning}}, model, \textcolor{blue}{\textbf{system}} \\

2019 & \textcolor{blue}{\textbf{data}}, \textcolor{blue}{\textbf{using}}, \textcolor{blue}{\textbf{based}}, model, \textcolor{blue}{\textbf{method}}, \textcolor{blue}{\textbf{algorithms}}, \textcolor{blue}{\textbf{techniques}}, \textcolor{blue}{\textbf{learning}}, \textcolor{blue}{\textbf{methods}}, models, \textcolor{blue}{\textbf{used}}, algorithm, ai, \textcolor{blue}{\textbf{system}}, field, ml \\
 
\end{tabular}
\caption{\abstracts{}}  \vskip -1.5ex
\label{tab: timeline abstracts}
\end{subtable}

\caption{Timelines of top 15 most frequently co-occuring words with ``artificial intelligence`` or ``machine learning'' within a window size of 5 in \articles{} and \abstracts{}. Words bolded in blue appear in all three years for each dataset respectively.}
\end{table*}

\subsection{Mutual Information: Strength of Association}
We measure the strength of association for words co-occurring with \textit{artificial intelligence} and \textit{machine learning} using mutual information (described in Section 3.2). We compute mutual information for words that have a relative frequency of at least 0.1\% for each year respectively to limit our analysis to popular words. Table \ref{tab:mi_table} presents mutual information (MI) and relative frequency (Frq) for the top five co-occurring words with the highest mutual information value over time in \articles{} and \abstracts{}. 

\begin{table*}[t!]\small
        \centering
        \begin{tabular}{c lrl  lrl  lrl}
        \toprule
         &\multicolumn{3}{c}{\textbf{2011}} & \multicolumn{3}{c}{\textbf{2015}} & \multicolumn{3}{c}{\textbf{2019}}\\ \cmidrule(lr){2-4} \cmidrule(lr){5-7} \cmidrule(lr){8-10}
         & \textbf{Word} & \textbf{MI} & \textbf{Frq} & \textbf{Word} & \textbf{MI} & \textbf{Frq} & \textbf{Word} & \textbf{MI} & \textbf{Frq} \\ \cmidrule(lr){2-4} \cmidrule(lr){5-7} \cmidrule(lr){8-10}
        
        \multirow{5}{*}{\textbf{\articles{}}} & mit's & 13.6 & 0.001 &  ai & 11.7 & 0.007 & ai & 10.9 & 0.02\\
        \rowcolor{Gray} & robotics & 12.9 & 0.004 & algorithms & 11.6 & 0.004 & algorithms & 10.6 & 0.003\\
        & algorithms & 12.5 &  0.004 & robotics & 11.5 & 0.004 & robotics & 10.5 & 0.004\\
        \rowcolor{Gray} & siri & 12.1 & 0.002 & azure & 10.8 & 0.001 & artificial & 9.8 & 0.007\\
        & ai & 11.9 & 0.006 & predictive & 10.6 & 0.002 & augmented & 9.8 & 0.001\\
        \\

        \multirow{5}{*}{\textbf{\abstracts{}}} & uci & 12.6 & 0.002 & uci & 12.5 & 0.002 & ai & 10.3 & 0.007  \\
        \rowcolor{Gray} & supervised & 10.8 & 0.002 & supervised & 10.9 & 0.003 & supervised & 9.7 & 0.002\\
        & ai & 10.7 & 0.005 & repository & 10.7 & 0.002 & classifiers & 9.1 &  0.001\\
        \rowcolor{Gray} & repository & 10.3 & 0.002 & ai & 10.4 & 0.003 & unsupervised & 8.9 & 0.001 \\
        & classifiers & 9.8 & 0.001 & classifiers & 9.9 & 0.002& algorithms & 8.9 & 0.009\\
        
        \bottomrule
        \end{tabular}
        \caption{Top five words with the highest mutual information to AI/ML over three years for \articles{} and \abstracts{}.}
        \label{tab:mi_table}
\end{table*}

\textbf{\articles{}:} We find that \textit{ai}, \textit{algorithms}, and \textit{robotics} consistently appear in the top five words with highest MI values, indicating that these words have a consistent and strong association to the terms \textit{artificial intelligence} and \textit{machine learning} in news articles. Interestingly, \textit{siri} and \textit{azure} appear in Table \ref{tab: timeline}, highlighting that these systems are disproportionately represented in the context of AI/ML in the news article corpus. Other words with high mutual information to \textit{artificial intelligence} and \textit{machine learning} change from year to year, with words like \textit{mit} and \textit{stanford} dropping in mutual information from the 2011 results to the 2019 results and words like \textit{blockchain} and \textit{cybersecurity} appearing first in the 2019 results. 

\textbf{\abstracts{}:} We find that \textit{ai}, \textit{classifiers}, and \textit{supervised} consistently appear in the top five words with highest mutual information values, indicating that these words have a strong association to the terms \textit{artificial intelligence} and \textit{machine learning} in \abstracts{}. The words \textit{uci} and \textit{repository} reference the UCI Machine Learning Repository, a data repository which currently warehouses 559 datasets for machine learning.\footnote{\url{https://archive.ics.uci.edu/ml/index.php}} None of the words that appear in Table \ref{tab: timeline abstracts} consistently over time appear in the Table \ref{tab:mi_table} consistently over time. While \textit{algorithms} appear in 2019 in Table \ref{tab:mi_table}, the rest of the words that have the highest co-occurrence frequency with \textit{artificial intelligence} and \textit{machine learning} are not distinctly unique to AI/ML. In general, words with high mutual information to \textit{artificial intelligence} and \textit{machine learning} remain consistent over time; however, few words have increasing mutual information, like \textit{deep} and \textit{big}.

\subsection{Normalized Co-occurrence Rank: Language Shifts Over Time}
We measure the shift of words co-occurring with AI/ML by computing the standard deviation of the normalized co-occurrence ranks for words with frequencies in the top 1\% of \articles{} and \abstracts{} for at least one year. This measurement produces 921 results for \articles{} and 457 results for \abstracts{}. Standard deviation values fall between 0 (no variation) and 0.47 (maximum variation) using this 1\% frequency threshold. Table \ref{tab: norm rank} displays results for the standard deviation values of 0, 0.05-0.1, 0.1-0.4, and 0.4-0.47 (limited to 20 words per bin) to showcase words with the least and the most variance over time (see Supplementary Materials for full results). For the words with fluctuating co-occurrence ranks, we examine the direction of their shift (decreasing in rank or increasing in rank), and if a word is not observed in 2011 but is observed in 2015 and 2019, we consider the word to be emerging.

\newcolumntype{R}{>{\raggedleft\arraybackslash}p{16mm}}
\newcolumntype{L}{>{\raggedright\arraybackslash}p{68mm}}
\newcolumntype{A}{>{\raggedright\arraybackslash}p{80mm}}

\begin{table*}\small
\centering
\begin{tabular}{R| AL}
\toprule
     \textbf{Rate of Change} & \textbf{\articles{}} & \textbf{\abstracts{}}  \\
     \midrule
     No shift \tiny$\sigma = 0$ & advanced, algorithms, computer, data, human, information, institute, language, mining, processing, research, researchers, robotics, science, software, system, techniques, technologies, university, use & analysis, classification, computational, data, engineering, information, methods, mining, model, network, neural, processing, recognition, repository, researchers, statistical, svm, technique, theory, used \\
     %\midrule
     %0.001-0.01 & brain, cloud, complex, company, engineering, expert, ibm, identify, intelligent, international, laboratory, learning, machine, mit, mobile, natural, networks, potential, predictive, recognition, speech, virtual, vision, voice, watson & clustering, cognitive, complex, database, dataset, detection, extraction, features, fuzzy, genetic, human, hybrid, logic, medical, regression, robotics, simulation, state-of-the-art, software, supervised, tool, uci, unsupervised, vision \\
     %\midrule
     %0.01 - 0.05 & analytics, automated, biotechnology, cognitive, computational, environment, financial, global, google, impact, innovative, integrated, manufacturing, medicine, military, neural, pioneer, possible, powerful, security, siri, sophisticated, space, vehicle, wireless & advances, attacks, autonomous, behavior, brain, clinical, cloud, data-driven, emerging, ensemble, forest, health, images, integrated, linear, mobile, modern, monitoring, real-time, risk, robot, robust, sensors, server, smart \\
     \midrule
     Minimal increase \tiny$\sigma \in [0.05, 0.1)$ & apps, capability, chips, competitive, cutting-edge, economic, education, government, investment, marketing, modern, monitoring, navigation, operational, quantum, revolution, risk, state-of-the-art, sensing, surveillance & adversarial, analytics, apparatus, deep, equipment, obtaining, operation, quantum, rapid, relates, storage, things, utility, vehicle, voice \\
     \midrule
     Significant increase \tiny$\sigma \in [0.1, 0.4)$ & cloud-based, defense, demand, drone, ethical, facebook, forecast, microsoft, nlp, novel, patent, policy, privacy, processors, rapid, saas, smartphones, stock, tesla, transforming & big, medium, terminal, unmanned\\
     \midrule
     Maximum increase \tiny$\sigma \in [0.4, 0.47]$ & apis, amazon, azure, bitcoin, blockchain, chatbots, commerce, cybersecurity, data-enabled, disruptive, ethereum, facial, flashstack, fintech, genomic, iot, newswire, selfdriving, semiconductor, startups & convolutional, discloses, iot \\
     \bottomrule

\end{tabular}
     \caption{Words (listed alphabetically) from \articles{} and \abstracts{} within binned normalized rank standard deviations. All words with $\sigma \in [0.4, 0.47]$ emerge in the 2015 subset of \articles{}/\abstracts{} (\textit{flashstack} emerges in the 2019 subset).}
     \label{tab: norm rank}
\end{table*}

\textbf{\articles{}:} Of the 921 resulting words from \articles{}, 17\% of words have $\sigma$ values in (0, 0.1], such as \textit{robotics} and \textit{software}, indicating a consistent co-occurrence frequency with AI/ML. Only two words (\textit{siri} and \textit{laboratory}) have downward trending co-occurrence ranks. Both words lose popularity from 2011 to 2015, but stay consistent from 2015 to 2019. The remaining words, such as \textit{ethical} and \textit{quantum}, have an upward trend in co-occurrence ranks. Emerging words, such as \textit{blockchain} and \textit{cybersecurity}, signal new application areas, systems, and products that are integrating AI/ML. Words with minimal increasing ranks not displayed in Table \ref{tab: norm rank} include company names and systems(e.g., \textit{ibm}, \textit{watson}, \textit{google}, \textit{siri}, and \textit{mit}) and application areas (e.g., \textit{biotechnology}, \textit{military}, and \textit{manufacturing}).

\textbf{\abstracts{}:} Of the 457 resulting words from \abstracts{}, 70\% of words have $\sigma$ values in (0, 0.1], such as \textit{theory} and \textit{statistical}, indicating a consistent co-occurrence frequency with AI/ML. Three words are labeled as emerging (\textit{convolutional}, \textit{discloses}, and \textit{iot}) and seven words (\textit{retrieval}, \textit{reasoning}, \textit{genetic}, \textit{web}, \textit{fuzzy}, \textit{cognitive}, and \textit{logic}) have minimally decreasing co-occurrence ranks. Words with increasing co-occurrence ranks signal new models, systems, and techniques (e.g., \textit{adversarial}, \textit{quantum}, and \textit{unmaned}).

\section{Discussion}
Generally, we find that the language surrounding AI/ML in news articles changes much more than in scientific publication abstracts. By measuring the strength of word associations and shifts in language over time, we find more consistent language use in \abstracts{} than in \articles{} (displayed in Table \ref{tab:mi_table} and \ref{tab: norm rank}). While frequently co-occurring words in \abstracts{} change minimally, frequently co-occurring words in \articles{} shift from words such as \textit{software} and \textit{research} to words like \textit{analytics} and \textit{digital}. 

%Our word association measurements provide insight into consistent, shifting, and emerging words frequently co-occurring with AI/ML. 

Our word association measurements provide insight into words that have a consistent, strong association to \textit{artificial intelligence} and \textit{machine learning}, as well words that have a shifting strength of association. Comparing mutual information values over time, we highlight words with strong associations to AI/ML over all three years. For example, in \articles{}, \textit{robotics} and \textit{robots} have consistently high mutual information values, aligning with Cave et al.'s finding that many adults define AI in relation to robots. Words with consistently high mutual information values in \abstracts{} identify commonly used models and fundamental components of AI/ML, such as \textit{supervised} \textit{repository}, and \textit{mining}. Mutual information results from \abstracts{} align with Krafft et al.'s finding that most AI researchers define AI in terms of its capabilities and applications in technical problems \cite{krafft2020defining}. 

Computing the standard deviation of normalized co-occurrence ranks, we highlight words that are consistent, shifting (including the shift's direction), and emerging in text. In \articles{}, emerging words signal new application areas (e.g., \textit{blockchain} and \textit{bitcoin}) and words increasing in rank signal booming application areas or improved products (e.g., \textit{smartphones} and \textit{chatbots}). Notably, in \articles{}, \textit{ethical} emerges in 2015, an appearance consistent with reports on increasing concerns in policy and society surrounding the ethical implications of AI models \cite{fast2017,cave2019scary,chuan2019framing}. In \abstracts{}, emerging words (e.g., \textit{convolutional}) highlight emerging models in AI/ML, while words increasing in rank (e.g., \textit{quantum}) highlight growing AI/ML application areas. 

%as our measurements dynamically capture the shifts in language over time.
These comprehensive results indicate how culture and trends affect how AI/ML are perceived, applied, and defined in the context of news articles and scientific publication abstracts. We are able to identify words that are consistent over time (e.g., \textit{algorithms}, \textit{computers} and \textit{data}), thereby demonstrating word association norms. We can also identify emerging words---specifically companies, products, systems, models, and technologies that have strong associations to AI/ML (e.g., \textit{facebook}, \textit{quantum}, and \textit{semi-conductor})---providing insight into the evolution of AI.

\begin{comment}
Analyzing word associations with the terms \textit{artificial intelligence} and \textit{machine learning} in news articles and scientific publication abstracts provides insight into how the public encounters AI/ML and how subject-area experts apply AI/ML. 

The challenge to provide a universally-accepted definition of AI remains, but this work presents a large-scale approach in identifying words that frequently appear in the context of AI/ML, signalling the ways in which an individual may define or perceive the terms \textit{artificial intelligence} and \textit{machine learning}. 
\end{comment}

\section{Conclusion}

Artificial intelligence is challenging to study, as it is an emerging and rapidly evolving technology that is actively integrated into various domains. Our work implements an automated analytical approach to study the language surrounding AI/ML over time in order to highlight consistent, shifting, and emerging language. We use two large-scale datasets from news articles and scientific research publications, applying our approach in a domain reflecting public perception and a domain reflecting subject-area applications. Capturing word association norms with AI/ML (e.g., \textit{robotics} and \textit{algorithms}), as well as emerging word associations (e.g., \textit{ethical} and \textit{cybersecurity}), our results not only align with prior manual research and surveys but also provide new insights into public perceptions and subject-area discussions of AI.

Interesting extensions of our analysis would be to use text corpora from different domains, such as social media text and policy documents as well as text in non-English languages, to provide a global perspective of AI.

\bibliographystyle{named}
\bibliography{ijcai21}

\end{document}